\documentclass[letterpaper, twocolumn, 11pt]{ieeetrans}
\newcommand{\linenumbers}{}

\usepackage{amsmath, amssymb, bm}

\usepackage[utf8]{inputenc}
\usepackage{graphicx}
\usepackage{xcolor}
\usepackage{accents}
\usepackage{authblk}

\usepackage[
  hypertexnames,%
  citecolor=blue,%
  colorlinks=true,%
  linkcolor=red,%
  bookmarks=true
]{hyperref}

\usepackage[sorting=none]{biblatex}
\usepackage[shortlabels]{enumitem}

\begin{document}
\linenumbers
\newcommand{\groupderiv}[2][]{\accentset{\scriptstyle\circ}{#2}}
\newcommand{\bodyvel}{\groupderiv{g}}
\newcommand{\group}{G}
\newcommand{\pos}{g}
\newcommand{\conn}{A}
\newcommand{\metric}{M}
\newcommand{\base}{R}
\newcommand{\shape}{r}
\newcommand{\shapevel}{\dot{r}}
\newcommand{\torq}{\tau}
\newcommand{\ctrl}{a}
\newcommand{\pass}{p}
\newcommand{\template}{\theta}
\newcommand{\templatevel}{\dot{\theta}}
\newcommand{\pert}{\delta}
\newcommand{\pertvel}{\dot{\delta}}

\title{Data-Driven Geometric System Identification \\ for Shape-Underactuated Dissipative Systems}

\author[1,2]{Brian Bittner}
\author[3]{Ross L. Hatton}
\author[1,4]{Shai Revzen}
\affil[1]{Robotics Institute, University of Michigan, Ann Arbor, USA}
\affil[2]{Johns Hopkins University Applied Physics Lab, Laurel, MD, USA \\ brian.bittner@jhuapl.edu}
\affil[3]{Collaborative Robotics and Intelligent Systems (CoRIS) Institute \&
School of Mechanical, Industrial, and Manufacturing Engineering, Oregon State University, Corvallis, USA\\ ross.hatton@oregonstate.edu}
\affil[4]{Electrical Engineering and Computer Science Department  \& Ecology and Evolutionary Biology Department, University of Michigan, Ann Arbor, USA \\ shrevzen@umich.edu}
\maketitle

\begin{abstract}
Modeling system dynamics becomes challenging when the properties of individual system components cannot be directly measured, and often requires identification of properties from observed motion. In this paper, we show that systems whose movement is highly dissipative have features which provide an opportunity to more easily identify models and more quickly optimize motions than would be possible with general techniques.
Geometric mechanics provides means for reduction of the dynamics by environmental homogeneity, while the dissipative nature minimizes the role of second order (inertial) features in the dynamics.
Here we extend the tools of geometric system identification to ``Shape-Underactuated Dissipative Systems (SUDS)'' -- systems whose motions are more dissipative than inertial, but whose actuation is restricted to a subset of the body shape coordinates.

Many animal motions are SUDS, including micro-swimmers such as nematodes and flagellated bacteria, and granular locomotors such as snakes and lizards.
Many soft robots are also SUDS, particularly robots that incorporate highly damped series elastic actuators to reduce the rigidity of their interactions with their environments during locomotion and manipulation.


We motivate the use of SUDS models, and validate their ability to predict motion of a variety of simulated viscous swimming platforms.
For a large class of SUDS, we show how the shape velocity actuation inputs can be directly converted into torque inputs, suggesting that systems with soft pneumatic or dielectric elastomer actuators can be modeled with the tools presented.
Based on fundamental assumptions in the physics, we show how our model complexity scales linearly with the number of passive shape coordinates.
This scaling offers a large reduction on the number of trials needed to identify the system model from experimental data, and may reduce overfitting.
The sample efficiency of our method suggests its use in modeling, control, and optimization in robotics, and as a tool for the study of organismal motion in friction dominated regimes.
\end{abstract}

\section{Introduction}

Rigid, fully actuated mechanisms are the classic face of robotics.
The development of passive elements \cite{pratt1995series, rouse2014clutchable, kalouche2015modularity, ahmadi1997stable, saranli2001rhex} and soft actuators \cite{klute1999mckibben, seok2012meshworm, pelrine2000high, shian2015dielectric} offers the potential for breakthrough improvements for the design of future systems.
Passive elements have the potential to assist in designing mechanisms that are safer, cheaper, more energy efficient, and more resilient to impact damage.
Inspiration from and mimicry of biology has played a strong role in the way that passive elements have been integrated into new types of mechanical devices.
For example, \cite{tolley2014untethered} demonstrated a soft quadrupedal robot which, while slow, was highly resilient -- it could be run over by a car and experience no damage;
\cite{huang2018chasing} presented a set of soft robotic designs that could achieve biomimetically competitive speeds, yet maintain at least some damage resilient properties thanks to their compliant nature.
In \cite{bujard2021resonant}, a soft robot mimicking  squid propulsion provided a way to achieve high efficiency swimming.
In all these cases these design improvements typically come at the cost of precise control of the internal state of the system.
Both the degree of underactuation of internal state and the complexity of the dynamics of soft mechanisms  exacerbate this problem.

Early robotics research showed that a convenient way to add compliance to a mechanism is to add a spring in series with a motorized joint \cite{pratt1995series}.
The ``Series Elastic Actuator (SEA)'' has been introduced to humanoids \cite{radford2015valkyrie} and snake robots \cite{rollinson2014design} with the goals of providing compliant, torque controlled interaction with the environment and higher damage resilience.
The design advantages of SEAs come at the expense of high-bandwidth position control.
It becomes difficult to execute precise body-shape trajectories that would be possible in the fully actuated, otherwise identical, systems.
In robots with soft actuators, the shortcomings in position control are exacerbated by the sensitive nonlinear dynamics of pneumatic devices, dielectric elastomers, and other soft actuation techniques \cite{webster2010design, rus2015design}.
The challenges of precise fabrication and assembly make it difficult to reliably reproduce dynamical outputs across copies of these devices.
Some recent work demonstrated data-driven, model predictive control of a soft robot comprised of McKibben actuators using Koopman theory based global linearization \cite{bruder2019modeling}.
Such an approach is critically dependent on the success of the linearizing transformation that converts this problem to one amenable to linear model predictive control.
Unlike the global linearization attempted by that Koopman analysis, whose domain of validity is still poorly understood, the approach we present here includes guarantees that the local models we build of the dynamics are accurate and truly linear in the shape velocity, at least to the extent that our assumptions about the physics hold.
Our results focus on a class of friction dominated robots, where we demonstrate an algorithmic procedure producing a complete and concise representation of the dynamics, informed by physics and geometry.

For fully actuated dissipative systems, previous work has provided sample-efficient techniques to model locomotion systems with noisy shape control using cyclic behavioral data \cite{bittner2018geometrically, kvalheim2019gait}.
Seminal work by Shapere, Wilczek, Marsden, Kelly, Ostrowski, Bloch and others \cite{shapere1989geometry, kelly1995geometric, bloch1996nonholonomic, marsden1998symmetries, ostrowski1998geometric, cendra2001geometric} showed that the mechanics of locomotion can be refactored into a kinematic term (the mechanical connection of \cite{marsden1998symmetries}) and a momentum term.
At the limit of large friction, the momentum term becomes negligible relative to the kinetic term, and the body velocity thus becomes a function of shape and shape velocity.
Previous work has demonstrated that this class of models are amenable to system identification \cite{bittner2018geometrically}.
Further, with finite-but-large dissipation, the influence of momentum can be folded into a nonlinear correction to the connection, with only a small increase in the complexity of the model identification process \cite{kvalheim2019gait}.
Models for predicting the influence of shape input on body velocity can thus be built strictly from observation without any mechanical analysis specific to the system -- all that is needed is ``sufficiently rapid'' dissipation of momentum.

In \cite{bittner2018geometrically} and \cite{kvalheim2019gait} we modeled the dynamics within ``closed tube'' regions of the configuration space using data from noisy periodic gaits that cycle within those tubes.
A key building block for these models is the ability to extract a reliable average periodic behavior from an ensemble of trajectories and re-express the observed trajectories in terms of phase in this periodic behavior plus a perturbation.
A video documenting the modeling process can be found in the supplementary materials of \cite{bittner2018geometrically}.

By having the ability to build local models, we gained the ability to modify maneuvers to maximize some behavioral reward, through incrementally improving the maneuvers in each local model.
A surprising conclusion of \cite{bittner2018geometrically} was that friction dominated systems can learn relatively high DOF optimal maneuvers (e.g. for an eight-jointed system) with only 30 cycles of data.
We showed these methods to be robust to actuator noise in section 8.2 of \cite{bittner2018geometrically} and observation noise in section 5.3 of \cite{bittner2018ddacm}.
In the presence of high actuator noise, robots optimizing behavioral efficiency tended toward higher amplitude motion, possibly because those are likely more robust to the experienced actuator noise.

One of the important consequences of the previous work is that it demonstrated we can produce a sample efficient behavioral model for systems that are hard to simulate, and for which we do not have model parameters such as inertia or drag matrices.
Traditional robot simulations rely on the kinematic and mass matrix information of each sub-system, but there are many practical scenarios where it would be preferable not to need this information.
For the data driven algorithms in \cite{bittner2018geometrically} and \cite{kvalheim2019gait}, we do not require such detailed mechanical information; we can select the appropriate algorithm by understanding (at a high level) how the friction and inertia interact in the system.

In the current work, we extend these ideas to underactuated systems.
First, we identify the class of ``Shape-Underactuated Dissipative Systems (SUDS)'' (see \S\ref{sec:models}) to which our methods apply.
Informally, these are systems that have fewer actuators than internal degrees of freedom and whose mechanics are governed primarily by frictional and damping forces, rather than by inertial forces.
We assert that SUDS are a highly useful and broad class of dynamical systems in practice.
We then show how data-driven geometric modeling techniques can be extended and used to identify predictive models for SUDS (see \S\ref{sec:models}).
For the subclass of SUDS whose internal dissipation is linear, the technique further allows us to collapse our model complexity, achieving a complexity that grows linearly in the degree of underactuation (see \S\ref{sec:est}).
To demonstrate the efficacy of our approach, we examine its performance on simulated viscous swimming data (see \S\ref{sec:eg}), validating that predictive SUDS models can be identified for soft, high dimensional systems with small amounts of trial data.
Finally, we discuss the relevance of SUDS identification in modern robotics applications.

\section{Background: Data-Driven Connection Modeling}
\label{sec:bg-ddgm}

In the field of geometric mechanics, the equations of motion arise from dynamical constraints derived from Lagrangian or Hamiltonian descriptions, after which group symmetries are applied to generate a reduced form of the dynamics~\cite{cendra2001geometric, bloch1996nonholonomic}.
The representation of these equations incorporates the uniformity of the operating environment and is achieved by
 ``quotienting the dynamics by symmetry under a group,'' i.e. exploiting the environmental homogeneity to re-write the equations of motion only in terms of body velocities and shape, without any dependence on absolute position and absolute orientation.

A common and representative case of group reduction is based on the symmetry that a body's interactions with a uniform environment do not depend on its position and orientation in that environment.\footnote{%
While our work applies without modification to other Lie group symmetries, we will tacitly assume that the symmetry is a subgroup of $\mathsf{SE}(3)$ and use the terms ``body frame'' and ``body shape'' for the ``fibre'' and ``base space projection'' that appear in the fibre bundle formulation of this theory.}
Under these circumstances we can re-write the equations of motion using a ``reconstruction equation''\cite{ostrowski1998geometric}, which appears as
\begin{align}
\bodyvel &= \conn(\shape)\shapevel + \mathbb{I}^{-1}(r) p \label{eq:gcircgeneral}
\\
\dot p &= f(r,\dot r, p)\label{eqn:reducedlagrangian}
\end{align}
where $\bodyvel$ is a velocity in the body frame, $\shape$ is an internal shape, and $p$ is momentum in the body frame.
These tools express in a formal and complete way the intuition that symmetry in the environment should allow us to write equations of motion relative to the body frame, and that shape changes can result in body motions either by directly pushing on the environment or by affecting the system's momentum.


Assuming viscous friction, as expressed in Lagrangian mechanics using a Rayleigh dissipation function, can further simplify behavior of the system.
At the upper limit of this friction, it is long known that a ``viscous connection'' emerges \cite{kelly1996geometry}. %
Rayleigh dissipation captures forces that are proportional to velocity.
Here, body velocity and momentum are both functions of shape and shape velocity, but neither is determined as a function of the other.
This connection governing the relationship between body and shape motion can arise as a result of constraints as well.
We can describe motion as being governed by linear constraints on the velocity; these are sometimes known as ``Pfaffian constraints'', and also result in a connection-governed system.
For moving systems with environmental symmetries, Pfaffian constraints often come in the form of body frame velocity constraints (e.g., no sideways slipping).
These systems are ``principally kinematic'' in the sense that their motion depends only on the path of their body configuration curve, but not on the rate.

The most well known, principally kinematic locomotors are viscous swimmers acting in low Reynolds environments \cite{hatton2013geometric}.
By exploiting the structure of the connection, tools have been developed for coordinate system selection, gait identification, and behavioral optimization \cite{hatton2011geometric, was2013optimal, hatton2015nonconservativity, wiezel2016using, ramasamy2019geometry}.

Predictive global models are often challenging to obtain for real animals and for physical hardware.
System identification techniques \cite{hatton2013geometric, dai2016geometric, schiebel2019mechanical, astley2020surprising} allow for data-driven modeling of animals and robots but require a large amount of experimental data.
Typically some reduction of the representation of the shape space is needed to make these methods produce tractable models of complex animals and robots.
Thus, there is a real need for modeling techniques with lean data requirements that can handle high dimensional representations of the body shape.

In \cite{bittner2018geometrically}, previous work developed a data-driven approach to geometric modeling and optimization.
It allows for the identification of a connection that governs a rhythmic motion with very little data (e.g. on the order of 30 cycles for a nine-link Purcell swimmer).
This estimation framework was built by combining oscillator theory \cite{revzen2009neuromechanical, revzen2008estimating, revzen2015data} and geometric gait optimization \cite{ramasamy2017geometric, ramasamy2019geometry}.
Using a phase estimator from \cite{revzen2008estimating}, phase is computed from observed cyclic shape data.
Grouping measurements by phase allows for the computation of a Taylor series approximation of the connection at each phase using linear regression across data gathered from multiple cycles.
Further theoretical analysis showed that when momentum decays quickly but not instantly, there exists a nonlinear $\conn(\shape,\shapevel)$ close to the linear connection; this additional nonlinearity was straightforward to capture with the inclusion of additional terms of the order of the momentum decay time-constant \cite{kvalheim2019gait}.

\section{Shape-Underactuated Dissipative Systems (SUDS)}
\label{sec:models}

The locomotion model for systems whose dynamics have the structure of a connection take the form
\begin{equation}
\bodyvel = \conn(\shape)\shapevel,
\label{eq:conn}
\end{equation}
where $r \in \mathbb{R}^n$ spans the shape space $R$, $g$ is an element of a Lie group $G$, and $A(r)$ is an infinitesimal lift from shape velocities to body velocities.
The notation $\groupderiv{g}$ denotes the world velocity $\dot{g}$ written in the body frame, computed as $g^{-1}\dot{g}$ for matrix Lie groups.

As discussed in \cite{hatton2013geometric, ramasamy2019geometry}, the internal wrenches along the shape degrees of freedom for a viscous-drag-dominated system can be written as:
\begin{equation}
\label{eq:metric}
\tau = -\metric(r)\dot{r},
\end{equation}
where $\metric$ is a Riemannian metric on the shape space generated by pulling back the local drag on individual portions of the system through its kinematics and the local connection.
Because Riemannian metrics are positive definite, the negation in equation \ref{eq:metric} implies that $\tau \cdot \dot r \leq 0$, i.e. that the system is dissipative and that changing shape always consumes energy.\footnote{
In control theory this property is referred to as being ``passive''.
}.

For underactuated systems, arbitrary choice of instantaneous shape velocity $\shapevel$ is infeasible.
Consequently, the form of equation \ref{eq:conn} is not directly useful for planning system motions.
To reformulate equation \ref{eq:conn} in a more useful structure, we split the shape configuration and force vectors as
\begin{equation}
\label{eq:split}
\shape = \shape_\ctrl \oplus \shape_\pass \qquad \tau = \tau_\ctrl \oplus \tau_\pass
\end{equation}
where $\ctrl$ indicates controlled degrees of freedom and $\pass$ indicates passive degrees of freedom.
These passive degrees of freedom are governed by some dynamical relationship in which the wrench on the passive joint is a function of shape, shape velocity, and body velocity,
\begin{equation}
\label{eq:passive-torque-dyn}
\tau_\pass = \tilde{f}(\shape,\shapevel,\bodyvel).
\end{equation}
We substitute equation \ref{eq:conn} into equation \ref{eq:passive-torque-dyn} to reduce this relationship to a mapping from shape and shape velocity to the internal wrenches on passive joints
\begin{equation}
\tau_\pass = f(\shape,\shapevel).
\label{eq:passive-elements}
\end{equation}

Following the same reasoning as in~\cite{ramasamy2020optimal}, we can use the $\ctrl\oplus\pass$ splittings of $\shape$ and $\tau$ to break $\metric$ into four blocks,
\begin{equation}
\metric = \begin{bmatrix} \metric_{\ctrl\ctrl} && \metric_{\ctrl\pass} \\ \metric_{\pass\ctrl} && \metric_{\pass\pass}\end{bmatrix},
\end{equation}
where for brevity we suppress the dependence of $\metric$ on $\shape$.
We can then represent the passive wrenches in two ways, drawing from equations \ref{eq:metric} and \ref{eq:passive-elements}, such that
\begin{equation}
\tau_\pass = -\metric_{\pass\ctrl}\shapevel_\ctrl - \metric_{\pass\pass}\shapevel_\pass = f(\shape,\shapevel),
\end{equation}
and after rearranging,
\begin{equation}
-\metric_{\pass\pass}\shapevel_\pass = f(\shape,\shapevel)+\metric_{\pass\ctrl}\shapevel_\ctrl.
\label{eq:passive-elements-with-metric-pieces-2}
\end{equation}

Noting that many physical systems, such as animals and robots, exhibit linear or nearly linear dissipation, we add the assumption that we may rewrite $f$, the wrenches applied to the passive elements, as an $\shape$ dependent affine (linear plus a constant) function of $\shapevel$,
\begin{equation}
f(\shape,\shapevel) = f_o(\shape) + F(r)\shapevel = f_o + F_\ctrl \shapevel_\ctrl + F_\pass \shapevel_\pass,
\label{eq:assumption}
\end{equation}
where we again suppress the dependencies of $f_o$ and $F$ on shape for brevity of notation.
Combined with equation \ref{eq:passive-elements-with-metric-pieces-2}, we arrive at a force balance in which each term is constant or linear in shape velocity,
\begin{equation}
-\metric_{\pass\pass} \shapevel_\pass = f_o + F_\ctrl  \shapevel_\ctrl + F_\pass \shapevel_\pass + \metric_{\pass\ctrl} \shapevel_\ctrl.
\end{equation}
Rearranging terms in this expression gives
\begin{equation}
-(\metric_{\pass\pass}+F_\pass) \shapevel_\pass = f_o + (F_\ctrl+ \metric_{\pass\ctrl}) \shapevel_\ctrl,
\end{equation}
demonstrating that that $\shapevel_\pass$ is affine in $\shapevel_\ctrl$.

Now we show that $(\metric_{\pass\pass} + F_\pass)$ is full rank, which will prove that the affine relationship between $\shapevel_\pass$ and $\shapevel_\ctrl$ is not degenerate.
Term $\metric_{\pass\pass}$ is positive definite as it is a diagonal block of $\metric$, which we have established is itself positive definite.
Term $F_\pass$ is semi-positive definite because any damped system will have a non-negative power dissipation from damping $\shapevel_\pass^T F_\pass \shapevel_\pass$.
The sum of a positive definite matrix and a semi-positive definite matrix is itself positive definite, and thus $(\metric_{\pass\pass} + F_\pass)$ is invertible.

Because equation \ref{eq:conn} is linear (and thus affine) in $\shapevel$, and $\shapevel_\pass$ is affine in $\shapevel_\ctrl$, we obtain that $\bodyvel$ must be affine in $\shapevel_\ctrl$.
The equations for ($\bodyvel$, $\shapevel_\pass$) are affine in $\shapevel_\ctrl$:
\begin{align}
\bodyvel &= \conn_\ctrl(\shape)\shapevel_\ctrl + \bodyvel_o(r) \label{eq:pred1} \\
\shapevel_\pass &= -(\metric_{\pass\pass}+F_\pass)^{-1}\begin{bmatrix} f_o + (F_\ctrl+ \metric_{\pass\ctrl}) \shapevel_\ctrl \end{bmatrix} \label{eq:pred2}
\end{align}

In many control applications the control input is $\tau_\ctrl$ rather than $\shapevel_\ctrl$.
Substituting \ref{eq:metric} into \ref{eq:pred2} provides an explicit affine formula relating $\tau_\ctrl$ to $\shapevel_\ctrl$
\begin{equation}
\tau_\ctrl = -\metric_{\ctrl\pass}\shapevel_\pass - \metric_{\ctrl\ctrl}\shapevel_\ctrl
\end{equation}

We define a ``Shape-Underactuated Dissipative System (SUDS)'' as a mechanical system operating within the dynamical constraints of equation \ref{eq:conn} and equation \ref{eq:metric}.
We focus on SUDS containing linear passive elements of the constrained form given by equation \ref{eq:assumption}.
These systems are therefore governed by motion models comprised of equations \ref{eq:pred1} and \ref{eq:pred2}.
When combined these equations lead to the observation that
\begin{equation}
(\bodyvel,\shapevel_\pass)^T = \tilde{C}(\shape) + B(\shape)\shapevel_\ctrl,
\end{equation}
i.e. the dynamics of SUDS are a nonlinear function of shape $\shape$, and are affine in the directly controlled shape velocity $\shapevel_\ctrl$.

\section{Estimation for SUDS}
\label{sec:est}

Now that we have established a dynamical characterization of SUDS, we can discuss the ramifications of this characterization for the estimation of motion models from data.
If analytical models are available, methods derived in \cite{ramasamy2020optimal} provide a way to perform gait optimization on drag dominated systems with an elastic joint.
However, when analytical models are not available, sample-efficient methods for system identification can greatly accelerate data-driven behavioral optimization.
We will show that the characterization presented in \S\ref{sec:models} will be important for data-efficient system identification of highly underactuated systems.
Following the approach described in our previous work \cite{bittner2018geometrically}, we focus on identifying the dynamics within a ``tube'' around a nominal (phase-averaged) trajectory $\template$ by expressing the shape as $\shape:=\template+\pert$.
We produce the nominal trajectory for a gait by logging the internal shape $r$ as a multivariate time series, obtaining an estimate of the asymptotic phase of every sample in the series, and taking the nominal trajectory $\theta$ to be a Fourier series average of the shape $r$ as a function of the phase.
Thus, the nominal trajectory is computed entirely from data.
The critical step of obtaining a good estimate of asymptotic phase is non-trivial; we used the \textit{phaser} algorithm \cite{revzen2008estimating}.

At the end of this process $\pert$ expresses deviation from the computed nominal trajectory $\template$.\footnote{%
  A convenient feature of this computed nominal trajectory is that it gets around the difficulty that the phase averaged trajectory of the internal state is hard to anticipate a-priori when a good model is not already present. %
  Here, the region of the dynamics we are estimating is adapted to be relevant in the space of observed behaviors of the system. %
}
We then consider the approximation of ($\bodyvel,\shapevel_\pass$) by a first-order Taylor expansion in ($\pert,\pertvel$) as
\begin{align}
(\bodyvel,\shapevel_\pass)^T \approx  &\tilde{C}(\template) + \frac{\partial \tilde{C}}{\partial \shape}(\template) \pert + B(\template)(\templatevel_\ctrl + \pertvel_\ctrl) \nonumber \\
&+ \frac{\partial B}{\partial \shape}(\template)\pert(\templatevel_\ctrl + \pertvel_\ctrl).
\end{align}
However, because $\templatevel$ is a predetermined function of $\template$, we can combine terms (suppressing the $(\template)$ for readability)
\begin{align}
C:= \tilde{C} +B\templatevel_\ctrl \\ C_\shape := \frac{\partial \tilde{C}}{\partial \shape} + \frac{\partial B}{\partial \shape} \templatevel_\ctrl
\end{align}
which provide the following linear regression problem at each $\template$,
\begin{equation}
\label{eq:reg}
(\bodyvel,\shapevel_\pass)^T \sim C + C_\shape \pert + B \pertvel_\ctrl + B_\shape \pert \pertvel_\ctrl.
\end{equation}

The regression in equation \ref{eq:reg} expresses the instantaneous body and shape velocities given the current shape (referenced from $\shape, \pert$) and the control input (referenced by $\pertvel_\ctrl$) to the system.

\subsection{SUDS balance compactness of model with capability to approximate dynamics}

A primary challenge in system identification is to select the model whose parameters will be extracted from the data.
Choosing too few parameters can cause underfitting while choosing too many parameters can often cause overfitting.
Here we show that the characterization of SUDS dynamics allows for a compact yet descriptive set of parameters to seed system identification.
In particular, we pay attention to the ability of the parameters to remain descriptive and concise at high degrees of underactuation, which is a prevalent feature of soft systems.
By highly ``descriptive'' we mean that the models maintain the capacity to explain the broad range of phenomena exhibited by SUDS.
By highly ``concise'' we mean that the number of regressors (also called ``features'' in machine learning) used to construct the model is small, allowing even small datasets to identify good models.

The overall shape space dimension is $n := n_\ctrl + n_\pass$, the number of directly controlled DoF and the number of passive DoF in the system respectively.
Compare now the regressors of equation \ref{eq:assumption} to those of a more general SUDS
\begin{enumerate}
    \item $\delta,\dot{\delta}$ for a first-order Taylor approximation of a general SUDS, having $O(n)$ unknowns.
    \item $\delta,\dot{\delta_\ctrl},\delta \otimes \dot{\delta}_\ctrl$ for a first-order Taylor approximation of a passive Stokesian system constrained as per equation \ref{eq:assumption}, having $O(n n_\ctrl)$ unknowns.
    \item $\delta,\dot{\delta},\delta \otimes \dot{\delta},\delta^2,\dot{\delta}^2$ for a second order Taylor approximation of the general SUDS, having $O(n^2)$ unknowns.
\end{enumerate}
Thus estimation (2) provides the structural context beyond (1) to accurately model system behavior while avoiding the $O(n^2)$ growth of estimation (3).
This has a clear advantage for soft systems, which typically have a small number of control inputs and a high dimensional shape space.
\section{Examples of SUDS Swimmers}\label{sec:eg}

To illustrate our method we examined several systems that are amenable to this estimation architecture.
In these systems, a viscous (``Stokes'') flow regime produced the affine constraints via Newtonian force balance.

\begin{figure*}
\includegraphics[width=\textwidth]{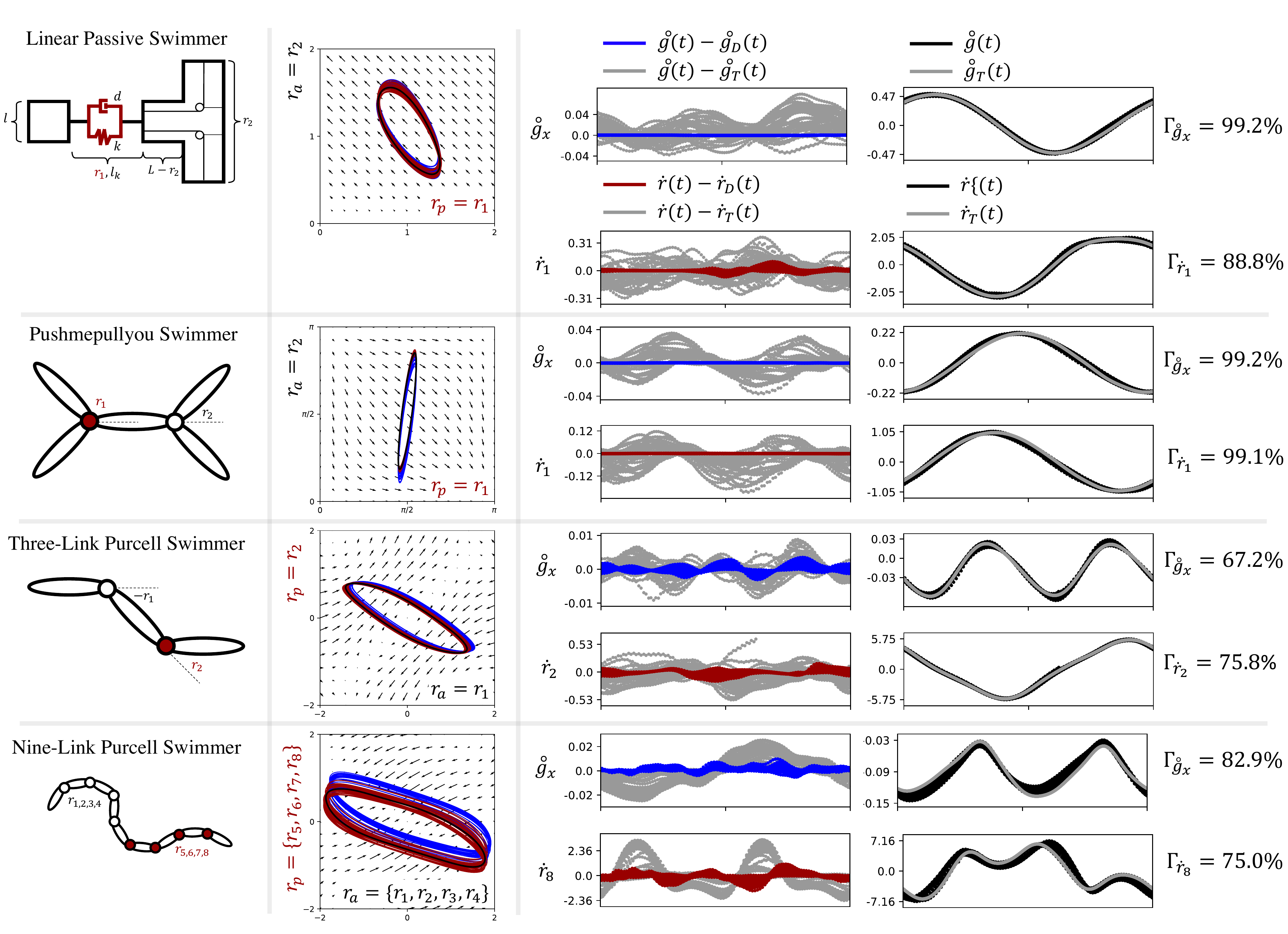}
\caption{Predictive quality of data-driven SUDS models for several systems. %
We examined the predictive ability of regressions in equation \ref{eq:reg} on simulated gait data for a linear passive swimmer, a pushmepullyou swimmer, a three-link Purcell swimmer, and a nine-link Purcell swimmer (top to bottom). %
The pulleys in the cartoon (first row of first column) indicate the constraints under which the bladder of the linear passive swimmer can deform.
In the cartoons of these systems (first column), we indicated controlled joints (black) and passive joints (red). %
We plotted the raw gait data (second column) including training data (red) and testing data (blue). We also include the phase-averaged gait of the training data (black) for each system.
We cycled each system at 1 Hz (such that $f=2\pi$ for each gait) for 30 cycles of training data and 30 cycles of testing data.
The metric $\Gamma$ provides a reference of how accurate the data-driven connection model is with respect to the phase averaged model. %
We compared the two models (third and fourth columns), plotting the residuals of data-driven body velocity model (blue) and passive shape velocity (red) on top of the phase averaged model residuals (grey). %
We also plotted passive shape and body velocity (black) with phase averaged model indicated (grey), demonstrating that while the phase averaged models are quite good, the data-driven connection model greatly improved the fidelity of the model, explained by the $\Gamma$ metric on the right.}
\label{fig:model-accuracy}
\end{figure*}

\subsection{Linear Passive Swimmer}
\label{sec:lps}
The linear passive swimmer (first row of Figure \ref{fig:model-accuracy}) consists of a shape-changing ``T-shaped'' paddle connected to a payload volume via a spring-damper.
The T shape is comprised of a horizontal bar of fixed width and variable length $\shape_2$, affixed to the midpoint of a vertical bar which has a fixed width and a dependent height $L-\shape_2$.
As $\shape_2$ varies, the faces of the paddle interact with a Low Reynolds fluid, generating reaction forces.
The spring-damper connection to the payload has rest length $l_k$, instantaneous length $\shape_1$, spring constant $k$, and internal damping coefficient $d$.
The viscous forces are modeled as a product of the length of the interacting face and the relative velocity of the face to the flow, scaled by a constant drag coefficient $c$.
Due to symmetry, the linear passive swimmer exerts no torques and it is constrained to move along the $x$ axis. The single Pfaffian constraint that describes the motion model is:
\begin{equation}
l\,c\dot{x} + c\shape_2(\dot{x}+{\shapevel_1}-{\shapevel_2}) = 0
\end{equation}
leading to the motion model
\begin{equation}
\bodyvel = \frac{-c\shape_2}{c(l+\shape_2)} \begin{bmatrix} 1 & -1 \end{bmatrix} \begin{bmatrix}{} \shapevel_1 \\ \shapevel_2 \end{bmatrix}{} \label{eq:pass-conn}
\end{equation}
(in which $\bodyvel = \dot{x}$).
This exact mechanical connection persists in the presence of shape-underactuation, which acts only to restrict what shape trajectories (and therefore group trajectories) can be expressed.
Note that we have not cancelled $c$ from the equation as a reminder of the fact that in more general cases the drag $c$ is a positive matrix rather than a scalar.

For this system, the internal forces can be written as
\begin{align}
lc\dot{x} &= k(\shape_1-l_k) + d\shapevel_1 + \omega \label{eq:internal-1} \\
c\shape_2(\dot{x}+\shapevel_1-\shapevel_2) &= k(l_k-\shape_1) -d\shapevel_1 + \omega \label{eq:internal-2},
\end{align}
where $\omega$ is the wrench that the world exerts on the system (in this case a force along the $x$-axis).

Combining the equations for external force balance (equation \ref{eq:pass-conn}) and internal force balance (equations \ref{eq:internal-1} and \ref{eq:internal-2}) provides three equations and three unknowns ($\bodyvel,\shapevel_3,\omega$).
We write the equations such that inversion of the matrix on the left-hand side will provide a locomotion model for the system's motion, given $\shape_1(t),\shape_2(t),\shape_3(t=0)$.
Stacking the equations, we write
\begin{align}
\begin{bmatrix} cl+c\shape_2 & 0 & c\shape_2 \\ cl & -1 & -d \\ c\shape_2 & -1 & (d+c\shape_2)   \end{bmatrix} \begin{bmatrix}{} \dot{x} \\ \omega \\ \shapevel_1 \end{bmatrix}{} = \nonumber \\ \begin{bmatrix}{} c\shape_2 \\ 0 \\ c\shape_2 \end{bmatrix}{} \shapevel_2 + \begin{bmatrix}{} 0 \\ k(\shape_1-l_k) \\ -k(\shape_1-l_k) \end{bmatrix}{}
\label{eq:linear-passive-swimmer}
\end{align}

The dynamics for the linear passive swimmer fit into the form of equations \ref{eq:pred1} and \ref{eq:pred2} where $\shape_\ctrl=\shape_2$ and $\shape_\pass = \shape_1$.
As a driving signal for this swimmer, we used $\shape_\ctrl := 1-\sin(ft)/2$, where f is a frequency parameter.
For physical constants, we used $L=2$, $l=0.5$, $c=1$, $d=0$, and $l_k=1$. While the internal damping coefficient in this and other platforms is set to zero, the viscous regime creates a damping-like resistance against internal body motions.

\subsection{Pushmepullyou Swimmer}
\label{sec:pmpy}
This symmetric viscous swimmer (second row of Figure \ref{fig:model-accuracy}), introduced in \cite{avron2008geometric} as the ``pushmepullyou swimmer," is constrained such that the pairs of links on the left and on the right open symmetrically about the center-line of the swimmer.
The symmetry allows us to assume the system moves only along the $x$ axis.
By exciting $\shape_1$ and making $\shape_2$ passive, we obtained a small forward displacement over every cycle.
We chose $L=1$, $k=10$, and $\shape_k=\frac{1}{2}\pi$.

The single Pfaffian constraint that drives the motion model is
\begin{align}
0 = L\dot{x} &+ 2(L c^2_1+2L s^2_1)\dot{x} + 2L^2  s_1 \shapevel_1 \nonumber \\
 &+ 2(L c^2_2+2L s^2_2)\dot{x} + 2L^2  s_2 -\shapevel_2
\end{align}
where for brevity, we denote $s_i, c_i := \sin(r_i), \cos(r_i)$ for $i= 1,2$.
This leads to the motion model
\begin{align}
\dot{x} = \alpha \begin{bmatrix} -L s_1 & L s_2 \end{bmatrix}^T \begin{bmatrix}{} {\shapevel}_1 \\ {\shapevel}_2 \end{bmatrix}{} = 0 \\
\alpha = \frac{1}{\frac{1}{2}+ c^2_1+2 s^2_1+ c^2_2+2 s^2_2}.
\end{align}
We place a spring on the left pair of joints such that $\shape_1$ is driven to $\shape_k = 0.5$rad via spring constant $k = 1$. We write the internal torque balance on the passive joint as
\begin{align}
k(\shape_1-\shape_0) = (-2L^2\shapevel_1+2L s_1\dot{x})L + \frac{L^3}{12}{\shapevel}_1.
\end{align}
This resulted in the equations
\begin{align}
\begin{bmatrix} \alpha^{-1} & L s_1 \\ \gamma_1 & \gamma_2 \end{bmatrix} \begin{bmatrix}{} \dot{x} \\ \shapevel_1 \end{bmatrix}{} =
\begin{bmatrix}{} L s_2 \\ 0 \end{bmatrix}{} -\shapevel_2 + \begin{bmatrix}{} 0 \\ k(\shape_1 - \shape_k) \end{bmatrix}{} \\
\gamma_1 =  2L^2 s_1 \hspace{1cm}
\gamma_2 =  -2L^3 + \frac{L^3}{12}
\end{align}
which match the form of equations \ref{eq:pred1} and \ref{eq:pred2}, where $\shape_\ctrl = \shape_2$ and $\shape_\pass = \shape_1$.
We drove this model with $\shape_\ctrl := \frac{1}{2}\pi+\frac{1}{3}\pi\sin(ft)$.

\subsection{Purcell Swimmer and nine-link viscous swimmer}
\label{sec:purcell}
The Purcell Swimmer and nine-link viscous swimmer (third and fourth rows of Figure \ref{fig:model-accuracy}) are known to have connection models \cite{avron2008geometric}.
Previous work \cite{bittner2018geometrically} studied the ability to model and optimize gaits with these platforms.
The force balance that induces the Pfaffian constraints is presented in \cite{hatton2013geometric}.
Torsional springs and dampers can act at the joints within the specified form of equation \ref{eq:assumption}, and the model will maintain the form of equations \ref{eq:pred1} and \ref{eq:pred2}.
In this work, we use the model and equations of \cite{hatton2013geometric}.
We use segment length $L=1$ with a spring at each passive joint having a rest angle of $0$.
For the three-link Purcell swimmer, we drove the controlled joint with $\shape_\ctrl := 1.4\sin(ft)$ and placed a spring with constant $k=2$ on the passive joint.
For the nine-link Purcell swimmer, we drove a traveling wave through the controlled joints $(1,2,3,4)$ with $\shape_{\ctrl,i} := 1.4\sin(f(t-i\phi))$ for $i \in [1,4]$ and $\phi=\frac{1}{4}\pi$.
We placed springs with constants $k = (20,15,10,5)$ for joints $(5,6,7,8)$ respectively.

\section{Estimator Accuracy} \label{sec:acc}

We sample the position and shape space of each of these systems at 100 time-steps per cycle for a 50 cycle trial.
The control inputs to the system were driven by a Stratonovich stochastic differential equation, in a process identical to that used in \cite{bittner2018geometrically}.
In summary, this process involves an input that is perturbed via Brownian noise while being exponentially attracted to a reference signal.
The reference is periodic, defining the gait or limit cycle that the system is perturbed about.
We select gaits for each system such that they noticeably excited the passive degrees of freedom.
We drive each gait at 1 Hz frequency as this was sufficient to produce excitation across all mechanisms. System length, frequency, and viscosity combine into Reynolds number, which can tell us how to adjust frequency to accommodate different physical parameters.
Likewise, this relationship can inform us how to change physical parameters to accommodate a required timing.
Choices such as the viscosity of the fluid and lengths of the swimmers can affect the timescales at which inputs excite the passive elements of the systems.
We compute each data-driven model by fitting the regressions equation \ref{eq:reg} to the trial data using the same method as \cite{bittner2018geometrically} (a fairly naive least squares regression approach).

To assess the quality of our data-driven models, we compare our SUDS regression models with the predictions obtained from a phase-averaged behavior of the same system.
Such phase-averaged behaviors can be viewed as the simplest ``template'' model of the dynamics, whereby all periodic locomotion gaits can be viewed as oscillators \cite{seipel2017conceptual}.
We employ the \textit{phaser} algorithm of \cite{revzen2008estimating} to reconstruct a phase from the ``observation'' data produced by the simulation, as this algorithm has been shown to be effective in producing phase driven models for many animal and robot locomotion systems \cite{maus2015constructing,wilshin2017longitudinal,council2020fast}.
Here, we denote by $\bodyvel$ and $\shapevel$, the ground truth body velocity and shape velocity samples (respectively). By $\bodyvel_T$ and $\shapevel_T$, we denote the predicted value for these quantities projected onto the phase model of the system
\footnote{%
  Equivalently, this can be considered a projection to the template system, which is a phase oscillator on the phase-averaged trajectory.%
}.
Finally, by $\bodyvel_D$ and $\shapevel_D$ we denote the data-driven model-predicted values of these same variables.

We define an accuracy metric for our predictions as one minus the ratio of the error in the data-driven prediction to the error in the phase-only predictions,
\begin{equation}
    \Gamma_* = 1 - \frac{\sum_{i=1}^m |*_D-*|}{\sum_{i=1}^m |*_T-*|},
\end{equation}
for $m$ samples and  $*=\{\bodyvel,\shapevel\}$.
$\Gamma_*=1$ indicates perfect prediction of the ground truth velocity, and $\Gamma_*=0$ means the model has no predictive improvement over using the phase-averaged behavior.
The data-driven models were notably more predictive than the template models, as illustrated in the right columns of Figure \ref{fig:model-accuracy}.

\section{Discussion and Conclusions}

We have shown that the broad class of ``Shape-Underactuated Dissipative Systems (SUDS)'' gives rise to dynamics that have an affine structure in the shape-velocity of their controlled DoF.
As a consequence, it was possible for us to formulate an efficient regression model of these dynamics and to demonstrate that for several simple models, these regressions would in fact improve prediction accuracy by a substantial factor.
Thus, we expanded on the capabilities of methods that can optimize analytical SUDS models \cite{ramasamy2020optimal} with methods that can fit SUDS models to data.
It is surprising that the first order Taylor approximation of a general SUDS can obtain an accurate model of the Purcell swimmer with just 30 cycles of 8-dimensional shape data.
This result is important for two reasons:
\begin{itemize}
  \item 8 dimensions is sufficient to capture fairly complex and broad ranges of internal geometries.
  \item The method is portable to systems that inhabit complicated environments that we have little chance of successfully modeling in simulation. The requirements are that the dynamics don’t change under translation and rotation of the body, momentum is rapidly dissipated, the process noise is not too large to build a reliable model, and the dynamics are not nonlinear in a way that the Taylor expansion of the SUDS model cannot describe the physics of the observed motion.
\end{itemize}
This result also suggests that the in the Purcell swimmer $A(r)$ is not very different from its first order Taylor approximation, at least for the gaits we explored.
This observation is consistent with our practical experience with simulated low-Reynlods number swimmers and with physical robots which use legs or undulatory motion, $\conn$ is only weakly nonlinear in $\delta$ for gaits $\theta$ which move the robot, i.e. approximating  by its first order Taylor expansion around $\theta$ works quite well.
However, even if the near-linearity we employ in our approximation broke down badly at some points on the cycle, as long as $\conn(\shape)$ remained bounded in value, the effects would be mitigated by the fact that the results go through a path-ordered integral.

The similarity of the results here to our previous work \cite{bittner2018geometrically, kvalheim2019gait} suggests that this would make it possible to rapidly learn behaviors in such underactuated systems.
It suggests that underactuation in SUDS does not pose nearly the same difficulties as in other underactuated systems --- the strong dissipation improves the stability of the passive dynamics under repeated but perturbed control inputs.

As is true of any attempt to experimentally determine equations of motion, the quality of the model we create is limited by the measurement noise and the amount of available data. Because we use system noise to identify a model of the $\delta$ and $\dot{\delta}$ dependence of $\conn(\theta+\pert)$, our method depends on the measurement noise being smaller than the system noise, and the system noise being small enough that a Taylor expansion of $\conn(\theta+\pert)$ in $\delta$ is still of good predictive value. One natural extension of our approach which could reduce its sensitivity to the scale of system noise is to use a more general non-linear function fitting method for capturing the $\delta$ dependence, e.g. Support Vector Regression (SVR), or any of the myriad other multivariate function fitting methods available currently or in the future. In this regard, our most important contribution is to note that much of the body velocity dependence on shape is actually affine due to the underlying physics, and thus easy to learn from data.

One particularly promising direction is modeling and control of soft systems with e.g. soft pneumatic actuators or systems with long, passive, flexible tails.
Many such biomimetic robots exist.
Earthworm-inspired robots \cite{joey2019earthworm} provide an opportunity to learn more about the engineering capabilities of our tools when applied to real hardware.
Applications to robots that move on varying surfaces, such as crawling on new branches \cite{rozen2019design}, may warrant methods that can build models with little data.
Analyses of simulation data for soft robotic behaviors like pipe navigation \cite{daltorio2013efficient} can be complemented or supported with information from behavioral models that can be generated from minutes of experimental data using our methods.

We have shown that our model identification regressions grow only linearly in complexity with the number of passive degrees of freedom.
Thus, we can reasonably hope to process high dimensional representations of the continuous (and thus ``infinite-dimensional'') shape of soft objects.
As long as the dimension of the representation provides a reliable state -- in the sense of having good enough predictive ability -- our work here provides good reason to believe the SUDS model identification will be tractable and produce predictive results.

From a biological perspective, we note that most animals are small (by human standards) and thus more dissipative because viscous friction (the phenomena which produces velocity proportional forces for the viscous swimmers) scales with area or length, whereas inertia scales with volume.
Futhermore, dynamic Coulomb friction can produce viscous-like properties, the analysis of which is a topic for future work.
The simplicity of SUDS modeling suggests that the control problem that small, and even more so small and aquatic, animals solve is thus fundamentally easier than the control problem faced by large terrestrial creatures such as ourselves.
We, therefore, offer the hypothesis that the neuromechanical control of animals is ancestrally geared for controlling SUDS and that the motor control ability of large-bodied extant species builds upon a more basal ability to learn to control SUDS.

A great part of the appeal of data-driven modeling to the robotics practitioner is the potential of our approach to systematically model the interactions of robots with unmodeled environments, even when these are potentially soft, compliant, and complex robots.
Because the model regressions are efficient and easy to update, one can envision online identification leading to a broadly applicable form of adaptive control.
This could allow robots to be highly adaptable to environmental changes and internal damage while retaining the ability to plan using the SUDS regression derived self-model.

The phase dependence of body velocity is relatively easy to estimate, but the linear dependence of the body velocity on changes in shape and shape velocity -- the terms linear in $\delta$, $\dot\delta$, and $\delta \otimes \dot \delta$  -- can more easily be obscured by noise in actuation and sensing.
To estimate these terms requires that the perturbations $\delta$ and $\dot\delta$ be larger than the intrinsic system noise.
The resulting estimate is, in a sense, a stochastic linearization of the underlying linear gains (see e.g. \cite{brahma2019quasilinear} for a modern treatment of stochastic linearization in control).

Having provided a generalized framework for modeling shape-underactuated dissipative systems from data, we hope to inspire implementations in locomotion, manipulation, and even biomedical devices.
For such applications, one needs to be sure that damping dominates inertia, and that the controlled subspace of the robot's shape is controlled in a responsive and accurate way.
Having these, the practitioner has access to a system identifier that is sample efficient enough to work in situ, i.e. in the actual operating environment, offering a broader space of practical applications for soft robots.
These could include planetary exploration, disaster scenarios with poorly characterized environments, and biomedical procedures.

\section*{Acknowledgement}
Revzen and Bittner thank NSF CMMI 1825918, ARO W911NF-17-1-0306, D. Dan and Betty Kahn Michigan-Israel Partnership for Research and Education Autonomous Systems Mega-Project, and AOFSR FA9550-20-1-0238 for funding.
Hatton thanks NSF Grant 1826446 for funding.

\printbibliography

\end{document}